\documentclass[letter, 10 pt, conference]{IEEEconf}  

\IEEEoverridecommandlockouts                              
\usepackage{lineno,hyperref}
\usepackage{xcolor}
\usepackage{subcaption}
\usepackage{graphicx}
\usepackage{mathtools}
\usepackage{afterpage}
\usepackage{tikz}
\usepackage{amsmath}
\usepackage{amsfonts}
\usepackage{siunitx}
\usetikzlibrary{patterns,decorations.pathreplacing}
\modulolinenumbers[5]

\newtheorem{remark}{\textbf{Remark}}

\begin{document}


\title{\LARGE \bf
  Data-driven Feedback Control of Lattice Structures \\ with Localized Actuation and Sensing 
}
%


\author{Dominik Fischer, Loi Do, Miana Smith, and Jiří Zemánek%
\thanks{This work was supported by the Grant Agency of the Czech Technical University in Prague, grant No. SGS22/166/OHK3/3T/13, and co-funded by the European Union under the project ROBOPROX (reg. No. CZ.02.01.01/00/22\_008/0004590)}
\thanks{Dominik Fischer, Loi Do, and Jiří Zemánek are with Faculty of Electrical Engineering, Czech Technical University in Prague
        {\tt\small \{fischdom, doloi, jiri.zemanek\}[at]fel.cvut.cz}. Miana Smith is with Center for Bits and Atoms, Massachusetts Institute of Technology, Cambridge, MA, USA}%
}





\maketitle
\thispagestyle{empty}
\pagestyle{empty}

\begin{abstract}

Assembling lattices from discrete building blocks enables the composition of large, heterogeneous, and easily reconfigurable objects with desirable mass-to-stiffness ratios. 
This type of building system may also be referred to as a digital material, as it is constituted from discrete, error-correcting components. 
Researchers have demonstrated various active structures and even robotic systems that take advantage of the reconfigurable, mass-efficient properties of discrete lattice structures.
However, the existing literature has predominantly used open-loop control strategies, limiting the performance of the presented systems. 
In this paper, we present a novel approach to feedback control of digital lattice structures, leveraging real-time measurements of the system dynamics.
We introduce an actuated voxel which constitutes a novel means for actuation of lattice structures.
Our control method is based on the Extended Dynamical Mode Decomposition algorithm in conjunction with the Linear Quadratic Regulator and the Koopman Model Predictive Control.
The key advantage of our approach lies in its purely data-driven nature, without the need for any prior knowledge of a system's structure. 
We illustrate the developed method via real experiments with custom-built flexible lattice beam, showing its ability to accomplish various tasks even with minimal sensing and actuation resources.
In particular, we address two problems: stabilization together with disturbance attenuation, and reference tracking. 
\end{abstract}



\section*{Notes}
This article has been accepted for presentation at the 2024 Conference on Decision and Control, taking place from December 16-19, 2024. 
When citing this work, please reference the version published in the official conference proceedings.


\section{Introduction}


Lattices are periodic structures that aim to achieve high stiffness and strength to weight ratios. 
Natural systems have demonstrated many examples of lattice structures, such as in honeycomb, leaf structure, or bone tissue. 
Because of their desirable mechanical properties, researchers across multiple domains, such as robotics, civil engineering, and architecture, have been studying these materials. 
However, manufacturing large objects with complex internal structures remains challenging. 
The decomposition of a period lattice into repeated discrete blocks, or voxels (volumetric pixel), can help address this as it enables the assembly of arbitrarily sized structures from the voxel feedstock. 
Such structures are also referred to as \textit{digital} structures, as they are composed of discrete elements instead of continuous matter.
Arranging building elements in repeating patterns to achieve unique and unconventional properties is also typical for metamaterials. 

A digital material system is inherently modular, enabling easier reconfigurability and replicability as compared to a continuously fabricated structure. 
Additionally, the voxelized system simplifies the processes of both making changes and repairs to the system since voxels may easily be switched out of the structure or added in. 
This substantially improves the repairability of the voxel system, as a failure in one part of the structure only requires the affected voxel to be removed, as opposed to replacing the entire structure, or adding a patch which might impact structure performance. 
By combining voxels with different mechanical structures, the resulting structure can have very specific bulk mechanical properties, including pre-programmed anisotropies, such as relative compliance in only one axis. 
This type of heterogeneous structure is difficult to achieve using conventional approaches to fabrication, in which either a combination of material types or significant design experience would be required to achieve this level of mechanical property control. 
Additionally, unlike standard approaches to fabrication, the size of the final structure is not constrained by the size of the machine used to make it. 
Instead, the digital material enables the assembly of almost arbitrarily sized structures. 
The discrete structure also allows for automated (dis)assembly of these materials. 


Digital structures have been predominantly studied in connection to their structural design~\cite{coulais_combinatorial_2016,gregg_ultra-light_2018, shaw_computationally_2019, tyburec_modular-topology_2022} and static behavior~\cite{jenett_meso-scale_2016}.
Active control of lattice structures has been previously explored for space applications, primarily for active damping, such as in~\cite{preumont_active_damping_1992} and comprehensive description of feed-forward and integral force feedback control of lattice-structured satellites was given in~\cite{paupitz_goncalves_dynamic_2007}. 
However, these works are not aimed at discrete lattice structures, limiting their extendibility. 
In recent years the focus has been shifting also towards studying and controlling the dynamic properties of metamaterials for larger deflections. 
In~\cite{parra_rubio_modular_2023}, the authors presented two structures composed by assembling voxel elements into a lattice: a morphing aircraft wing and an underwater snake-like robot. 
Active control was introduced into these structures allowing the wing to change shape and the robot to achieve swimming, nature-inspired motion, representing a modular approach to approximately continuum-style robotics.  
Among other fields where digital materials find relevance are soft robotics, production of wearable devices, prosthetics~\cite{mirzaali_shape-matching_2018}, as well as optimized electronic devices~\cite{langford_automated_2016}. 
Possible applications of digital metamaterials extend even to the construction of space structures~\cite{jenett_bill-e_2017, gregg_ultralight_strong_2024}, where autonomous structures with active shape control are of significant interest.

Our primary focus is on robotic structures akin to those presented in~\cite{parra_rubio_modular_2023}, where a voxel structure with specific anisotropies is actuated.
These structures have so far been subjected to precomputed input sequences, leaving space for more advanced control algorithms to be explored and applied. 
In this paper, we present a novel mechanism to actuate digital lattice structures and develop a systematic method for feedback control synthesis.
Controlling digital structures presents a challenge, primarily because it is difficult to obtain a dynamical model of a structure with many distinct building blocks. 
Furthermore, flexible structures naturally exhibit nonlinear dynamics.
In light of that, we opted for a data-driven modeling approach based on the Extended dynamic mode decomposition (EDMD) algorithm~\cite{williams_extending_2016}.
Using the EDMD, we were able to obtain a linear predictor, allowing us to use standard linear control synthesis methods.
We use Linear Quadratic Regulator (LQR) and Koopman Model Predictive Control (KMPC) to address the tasks of stabilization disturbance attenuation, and reference tracking.  
The process and results presented in this paper enable the solution of more complex tasks and expand the potential applications of digital lattice-based structures.



\section{System Description and Problem Definition}


\subsection{Cuboctahedron as Building Block}

We use a face-connected cuboctahedron lattice, i.e., a tiled polyhedron with eight triangular and six square faces.
In particular, we adopt a construction design developed by the \textit{Center for Bits and Atoms} at MIT published in~\cite{jenett_discretely_2020}.
A cuboctahedron voxel is assembled using six square faces connected at the corners. 
The faces are 3D-printed using PETG plastic on a commercial 3D printer, and are connected using rivets. 
Voxels are assembled together face-to-face with rivets.

\begin{figure} [tb]
  \centering
  \begin{subfigure}{.11\textwidth}
    \centering
    \includegraphics[width=\linewidth]{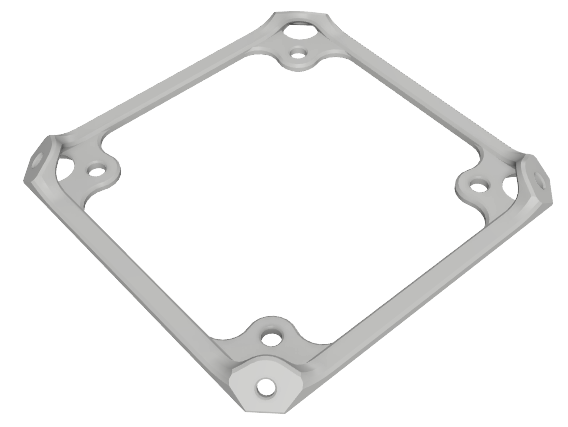}
    \caption{}\label{fig::rig_face}
  \end{subfigure}%
  \begin{subfigure}{.11\textwidth}
    \centering
    \includegraphics[width=\linewidth]{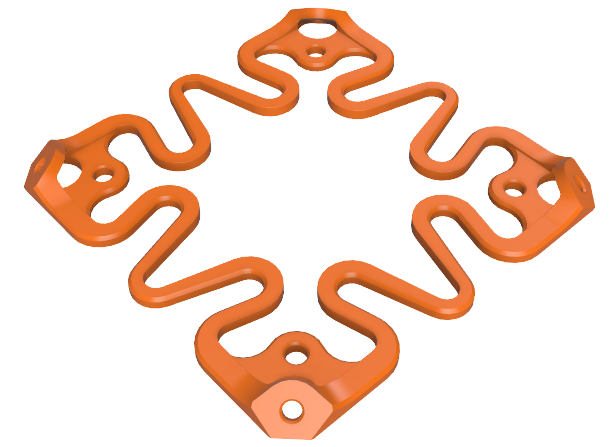}
    \caption{}\label{fig::cmp_face}
  \end{subfigure}
    \begin{subfigure}{.11\textwidth}
    \centering
    \includegraphics[width=\linewidth]{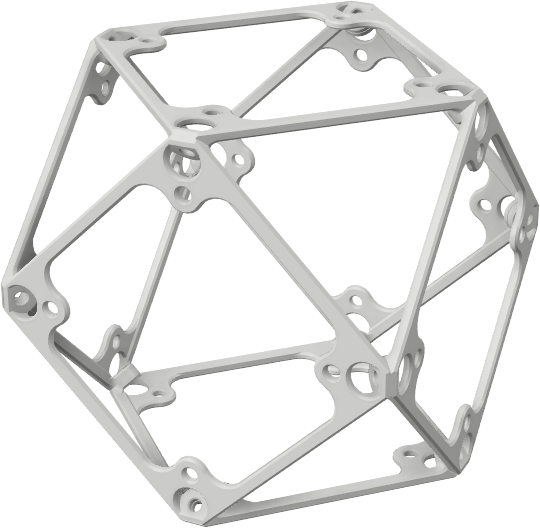}
    \caption{}\label{fig::rig_voxel}
  \end{subfigure}%
  \begin{subfigure}{.11\textwidth}
    \centering
    \includegraphics[width=\linewidth]{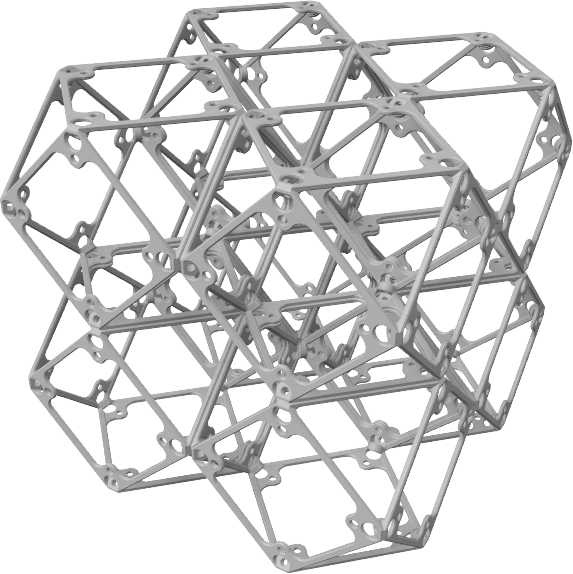}
    \caption{}\label{fig::rig_struct}
  \end{subfigure}
  \caption{(\subref{fig::rig_face}) The rigid face, (\subref{fig::cmp_face}) the compliant face, an assembled voxel (\subref{fig::rig_voxel}), and larger structure (\subref{fig::rig_struct})}\label{fig::BuildingBlocks}
\end{figure}

By changing the geometry of each face, we can compose cuboctahedra with different mechanical properties. 
In Fig.~\ref{fig::BuildingBlocks}, we show an example of two different face types, the assembled cuboctahedron, and a structure from multiple cuboctahedra.
Because the voxels are made from plastic, the assembled structures have the potential to be highly flexible with many degrees of freedom.

%

\subsection{Actuated Voxel}

\begin{figure} 
  \centering
  \begin{subfigure}{.20\textwidth}
    \centering
    \includegraphics[width=\linewidth]{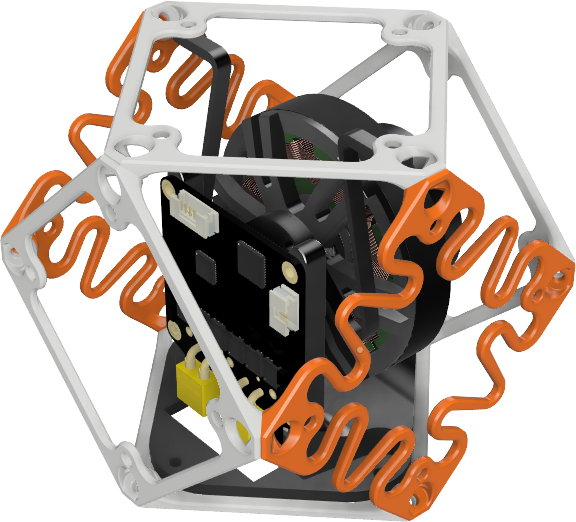}
    \caption{}
    \label{fig::act_vox}
  \end{subfigure}%
  \begin{subfigure}{.20\textwidth}
    \centering
    \includegraphics[width=\linewidth]{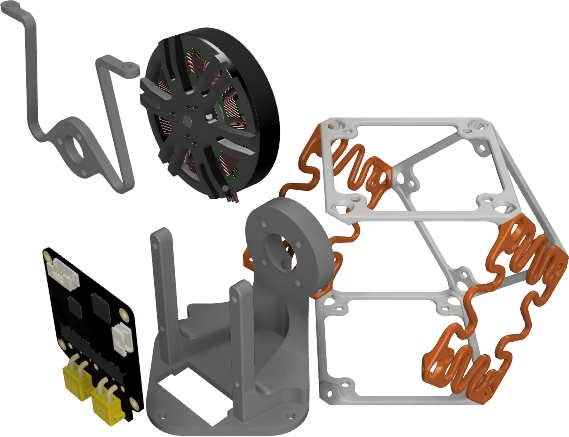}
    \caption{}
    \label{fig::act_vox_disas}
  \end{subfigure}
  \caption{Render of the actuated voxel: (\subref{fig::act_vox}) assembled; (\subref{fig::act_vox_disas}) individual parts} 
    \label{fig::actuated}
  \end{figure}

\begin{figure}
    \centering
      \begin{subfigure}{.15\textwidth}
      \centering
      \includegraphics[width=.97\linewidth]{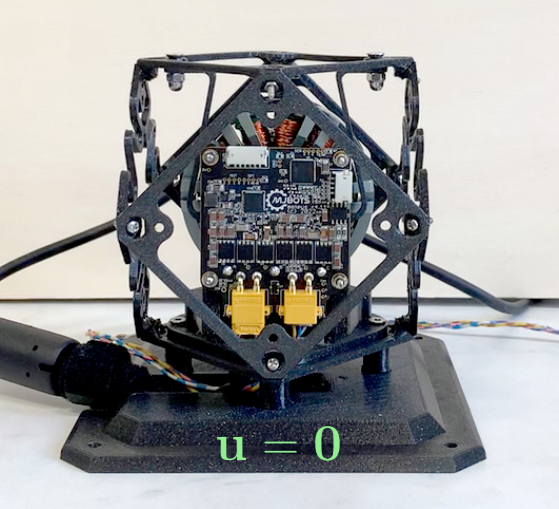}
      \caption{}
      \label{fig::AV_untilted}
      \end{subfigure}%
      \begin{subfigure}{.15\textwidth}
      \centering
      \includegraphics[width=.97\linewidth]{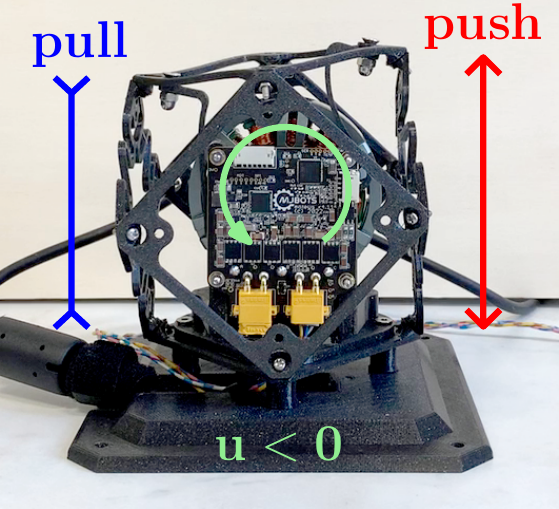}
      \caption{}
      \label{fig::AV_left}
      \end{subfigure}%
      \begin{subfigure}{.15\textwidth}
      \centering 
      \includegraphics[width=.97\linewidth]{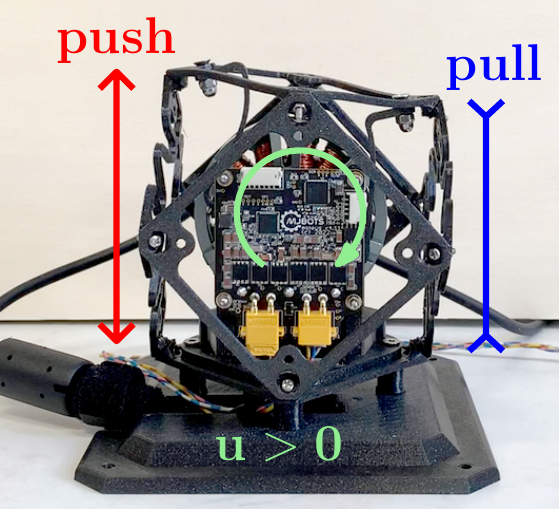}
      \caption{}
      \label{fig::AV_right}
      \end{subfigure}
      \caption{Demonstration of the real actuated voxel
      (\subref{fig::AV_untilted}) at rest;
      (\subref{fig::AV_left}) tilted to the left;
      (\subref{fig::AV_right}) tilted to the right}
      \label{fig::AV_real_movement}
  \end{figure}

We developed a novel mechanism for inducing controlled motion in the overall structure.
Compared to other actuation mechanisms for lattice structures (see Remark~\ref{rem:different_actuation}), our mechanism is contained within a voxel, making it fully compatible with the overall voxel assembly system.
The actuation is based on deforming a voxel with an internally housed motor. 
The actuated voxel is displayed in Fig.~\ref{fig::act_vox} with its individual components in Fig.~\ref{fig::act_vox_disas}.
The housing voxel is created from two compliant faces and four rigid faces in a configuration that allows significant deformation along one axis and relative stiffness in the others.
The deformation is induced by a motor that generates a torque which is then converted into a push-pull motion by a compliant transmitter that is connected to the top face of the voxel. This allows to directly deform specific places in the structure.
The motor is a \textit{mj5208} brushless DC motor controlled by the Moteus r4.11 driver by \textit{mjbots}, which allows desired torque to be set directly.
We illustrate the motion principle in Fig.~\ref{fig::AV_real_movement}, where we show three snapshots of the physical prototype.

\begin{figure}[tb]
    \centering
    \includegraphics[width=.8\linewidth]{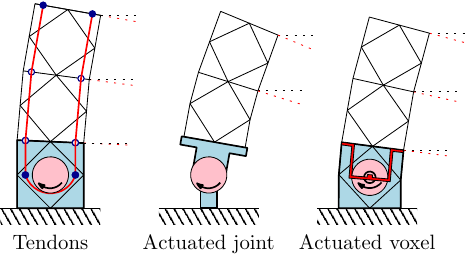}
    \caption{Comparison of actuator types}\label{fig::actuators_comparison}
\end{figure}

\begin{remark}\label{rem:different_actuation}
  One option to actuate voxel-lattice structures introduced in~\cite{parra_rubio_modular_2023,jenett_discretely_2020} is using \textit{tendons}, i.e., anchored wires, that span multiple voxels.
  The structure's motion is then induced by suitably shortening the tendons' lengths.
  Another option, frequently utilized in modular robotics, is to use a separate distinct module, an actuated \textit{joint}, such as in~\cite{smith_recursive_2023,abdel-rahman_self-replicating_2022}.
  In contrast, the actuation mechanism presented in this paper is fully contained in a single voxel.
  Additionally, unlike a specific component such as actutated joint, the housing voxel does not differ from other building blocks, which simplifies the construction of the overall structure.
  See Fig.~\ref{fig::actuators_comparison} for an illustration of differences between the mechanisms.
\end{remark}

\subsection{Sensing Voxel}

\begin{figure}[tb]
  \centering
  \includegraphics[width=0.8\linewidth]{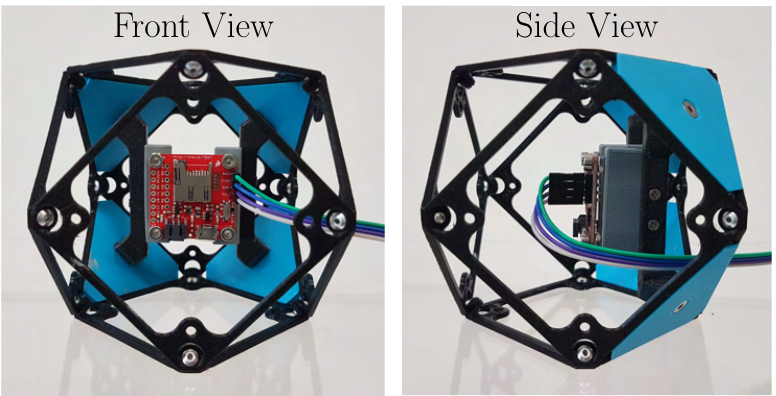}
  \caption{Sensing Voxel}\label{fig::sensing_voxel}
\end{figure}

For measuring the system's motion, we use an inertial measurement unit (IMU) inserted into a voxel. See Fig.~\ref{fig::sensing_voxel} for reference.
Similar to the actuated voxel, the sensing is also self-contained within a single voxel.
We created the housing voxel using only rigid faces, but different options could also be possible with minor changes to the construction.
We used an \textit{MPU9250} IMU, which allows measurement of acceleration, angular speed, and estimated orientation along all three axes.

\subsection{Problem Definition}

Consider a general (arbitrary) voxel-lattice structure with a fixed arrangement of empty, actuated, and sensing voxels.
We assume that the arrangement of the structure allows the input $u$ to control the output $y$.

Let the structures' dynamics be in a form of a discrete-time nonlinear system
\begin{subequations}\label{eq:general_nonlin_sys}
  \begin{align}
    x_{k+1} &= f(x_k, u_k) \;, \\ 
    y_k &= h(x_k) \;,
  \end{align}
\end{subequations} 
where $x \in \mathbb{R}^n$ is the system's state, $u \in \mathbb{R}^m$ is the control input, and $y_k \in \mathbb{R}^q$ is the system's measured output.
Both $f(\cdot)$ and $h(\cdot)$ are unknown, generally nonlinear functions.
The task is to design a feedback control for the system~\eqref{eq:general_nonlin_sys} to get a desirable behavior of the system's output $y$, e.g., disturbance attenuation or reference tracking.

\section{Linear Predictor based on Koopman Operator Theory}


To address the defined problem, we employ a data-driven approach. 
We start by identifying a \textit{predictor} for the dynamics~\eqref{eq:general_nonlin_sys}.
By predictor, we mean an artificial dynamical system designed to predict the original system's dynamics (or output) based on the initial state and the input sequence.
We restrict the predictor to have a linear system structure, so it can be then used for control synthesis through linear controller design methodologies.
The data-driven approach is particularly well suited for digital structures as it provides a systematic way to identify the system's dynamics, regardless of the system's structure and placement of controllers or sensors.
The identification is based only on supplying a suitable input signal into the system and measuring the outputs.


To obtain accurate predictions of the nonlinear dynamics using only a linear predictor, the main idea is to \textit{lift} the nonlinear dynamics into a higher-dimensional state-space.
With the original state $x \in \mathbb{R}^n$ of the nonlinear system, we consider the predictor in a form of a discrete-time linear system
\begin{equation}\label{eq:linear_predictor}
  \begin{split}
  z_{k+1} &= Az_{k} + Bu_k \;,\\
  \hat{y}_k &= Cz_k \;,
  \end{split}
\end{equation} 
with $\hat{y}_k$ being the prediction of the nonlinear system's output $y_k$, and $z_k \in \mathbb{R}^N, N \gg n$ being the \textit{lifted} state.
The concept of getting better prediction using lifting is supported by the \textit{Koopman operator theory} which we briefly explain in the next section.

However, from the practical point-of-view, the Koopman operator theory only provides theoretical foundations for the state-space \textit{lifting}.
The main challenge is finding the matrices $A,B$ and $C$ that represent the nonlinear dynamics well, for which we use the Extended Dynamic Mode Decomposition (EDMD) algorithm.




\subsection{Koopman Operator}

We provide only a brief introduction to Koopman operator theory, as our main goal is to establish the context for the terminology used.
For more rigorous exposition of Koopman operator theory, we refer the reader to~\cite{budisic_applied_2012} or~\cite{brunton_modern_2022}.
The foundations of the Koopman operator for analyzing dynamical systems were established by the works of Koopman and von Neumann in the 1930s~\cite{koopman_hamiltonian_1931,koopman_dynamical_1932}. 
We restrict ourselves to the case of autonomous systems.
The extension to systems with external inputs can be found in~\cite{korda_linear_2018}.

Consider an autonomous discrete dynamical system
\begin{equation}\label{eq::Koopman_autonomous}
x_{k+1} = \mathcal{T}\left(x_k\right),\qquad x_k \in \mathcal{M} \;, 
\end{equation}
where $\mathcal{T}(x_k)$ is a nonlinear transition mapping and $\mathcal{M}$ is a state space.
The core idea is to shift the focus from the mapping $\mathcal{T}$ to some (possibly infinitely many) user-defined functions of the states, so-called \textit{observables}. 
An observable is a function $\psi:\mathcal{M}\longrightarrow \mathbb{R}$ which belongs to typically an infinitely dimensional vector space $\mathcal{F}$.
The Koopman operator $\mathcal{K}: \mathcal{F} \longrightarrow \mathcal{F}$ is then defined as
\begin{equation}
  \left(\mathcal{K}\psi\right)\left(x_k\right) = \psi\left(\mathcal{T}(x_k)\right) = \psi\left(x_{k+1}\right)\;,
\end{equation}
Thus, the operator $\mathcal{K}$ advances every observable $\psi$ from the time step $k$ to $k+1$. 
One of the fundamental properties of $\mathcal{K}$ is its linearity since for any two observables $\psi^1$ and $\psi^2$ and scalars $\alpha$ and $\beta$ the following holds
\begin{equation}
  \mathcal{K}\left(\alpha\psi^1 + \beta\psi^2\right) = \alpha\mathcal{K}\left(\psi^1\right) + \beta \mathcal{K}\left(\psi^2 \right) \;.
\end{equation}
Therefore, using the Koopman operator, one can convert the analysis of finite-dimensional nonlinear system into the analysis of the infinite-dimensional linear operator.
Additionally, the Koopman operator fully captures the behavior of the nonlinear system, provided the space of observables contains the coordinate identity mappings.
Linear predictors in the form~\eqref{eq:linear_predictor} can then be viewed as finite-dimensional approximation to the Koopman operator.




\subsection{Linear Predictor Identification}\label{sec::DMD}


We identify the linear predictor~\eqref{eq:linear_predictor} of the system's dynamics~\eqref{eq:general_nonlin_sys} using the approach described in~\cite{korda_linear_2018}.
The approach is based on the Extended Dynamic Mode Decomposition (EDMD) algorithm for controlled systems.
The EDMD was first introduced in~\cite{williams_datadriven_2015}, and then generalized to the control setting in~\cite{korda_linear_2018}.
Full explanation and further information regarding the algorithm can be found in~\cite{korda_convergence_2018} or~\cite{brunton_modern_2022}.


The process starts with gathering $p$ data collections, the measurements and inputs $(y^{i}, y^{i}_{\texttt{+}}, u^i)$, $i = 1, \ldots, p$ from the nonlinear system satisfying
\begin{equation}
  \begin{split}
    y^i &= h(x_k) \;, \\
    y^{i}_\texttt{+} &= h\left(f(x_k, u^i)\right) \;, \\
  \end{split}
\end{equation}
where $h$ and $f$ are from~\eqref{eq:general_nonlin_sys}. Therefore, $y^i$ and $y^{i}_\texttt{+}$ are temporally ordered but $y^i$ and $y^{i+1}$ are necessarily not, so the measurements can be from different trajectories.
Then, we select $s$ lifting mappings $\psi^j$ -- the observables -- and combine all observables into a single vector function 
\begin{equation}
  \Psi(y) = \left[\psi^1\left(y\right), \psi^2\left(y\right), \cdots, \psi^s\left(y\right) \right]^\top \;.
\end{equation} 
We arrange the data collections in the following manner 
\begin{subequations}
  \begin{align}
    Y                       &= \left[ y^1, y^2, \ldots, y^p \right]  \;,\\
    Y_\texttt{+}            &= \left[ y^1_\texttt{+}, y^2_\texttt{+}, \ldots, y^p_\texttt{+} \right]  \;,\\
    Y^{\rm lift}            &= \left[ \Psi(y^1), \Psi(y^2), \ldots, \Psi(y^p) \right] \;, \\
    Y_\texttt{+}^{\rm lift} &= \left[ \Psi(y^1_\texttt{+}), \Psi(y^2_\texttt{+}), \ldots, \Psi(y^p_\texttt{+}) \right] \;, \\
    \Omega                  &= \left[ u^1, u^2, \ldots, u^p \right] \;.
  \end{align}
\end{subequations}
The linear predictor~\eqref{eq:linear_predictor} can be then obtained by solving 
\begin{subequations}\label{eq:EDMD_min}
  \begin{align}
    \min_{A,B}  &  {\lVert Y^\mathrm{lift}_\texttt{+} - A Y^\mathrm{lift} - B \Omega \rVert}_\mathrm{F} \;, \\
    \min_{C}    &  {\lVert Y - C Y^\mathrm{lift}            \rVert}_\mathrm{F} \;,
  \end{align}
\end{subequations}
where $ \lVert \cdot \rVert_\mathrm{F}$ is the Frobenius norm.
The analytical solution to~\eqref{eq:EDMD_min} is
\begin{equation}
  \begin{bmatrix}
    A & B \\
    C & 0 
  \end{bmatrix}
  =
  \begin{bmatrix}
    Y_\texttt{+}^{\rm lift} \\
    Y
  \end{bmatrix}        
  \begin{bmatrix}
    Y^{\rm lift} \\
    \Omega
  \end{bmatrix}^\top 
  \left(
  \begin{bmatrix}
    Y^{\rm lift} \\
    \Omega
  \end{bmatrix}
  \begin{bmatrix}
    Y^{\rm lift} \\
    \Omega
  \end{bmatrix}
  ^\top 
  \right)^\dagger     \;,
\end{equation}
where $(\cdot)^\dagger$ denotes the Moore--Penrose pseudoinverse~\cite{kutz_dynamic_2016}. 

\section{Control Methods} \label{sec::methods}

Having the linear predictor~\eqref{eq:linear_predictor}, we can design a feedback controller via classical linear control systems method.
We use two control techniques, the Linear Quadratic Regulator (LQR), and the Model Predictive Control (MPC).


\subsection{Linear Quadratic Regulator} \label{sec::LQR}
The discrete-time Linear Quadratic Regulator (LQR) is a controller that minimizes the quadratic cost function
\begin{equation}\label{eq::LQR_cost}
J(u,z) = \sum_{k=0}^{\infty} \left( z_k^\top Qz_k + u_k^\top R u_k \right) \;,  
\end{equation}
subjected to the dynamics of the system~\eqref{eq:linear_predictor}.
The matrices ${Q\succeq 0}$ and ${R\succ 0}$ represent the penalties on states and input, respectively. 
The control law minimizing~\eqref{eq::LQR_cost} is 
\begin{align}
    u_k = - Kz_k = -K \Psi(y_k) \;,
\end{align}
where $K$ can be computed from the solution to the discrete-time Algebraic Riccati Equation.


\subsection{Model Predictive Control}
Use of linear prediction in conjunction with model predictive control was first introduced in~\cite{korda_linear_2018}, and subsequently coined as \textit{Koopman} MPC (KMPC).
KMPC for control of dynamical systems has been previously used, such as in~\cite{do_controlled_2023,arbabi_data-driven_2018}.

Model Predictive Control (MPC) is a discrete-time optimal control algorithm that uses optimization to find input sequences that minimize a given cost function $J(\cdot)$ over defined time frame-- the \textit{prediction horizon} $N_\mathrm{p}$. 
Upon computing the optimal input sequence, the controller then applies only the first element from the sequence, and the process is repeated in the next time step.
With linear predictor of the system's dynamics, the optimization problem is a convex quadratic program for which many high-performance solvers exist. 
We use the OSQP solver~\cite{stellato_osqp_2020}.

We formulate the MPC for tracking.
Let $e_k$ be the tracking error and $r_k$ the reference.
At every time step $t$, the MPC solves an optimization problem
\begin{equation}\label{eq:MPC_basic_opt_problem}
\begin{split}
\min_{u_k,z_k}                \quad &   J\left(\{z_k\}_{k=0}^{N_\mathrm{p}}  \;, \{ u_k \}_{k=0}^{N_\mathrm{p}} \right)   \;, \\
\textnormal{subject to }      \quad &   z_{k+1} = Az_k + Bu_k\;,          \quad k=0,1,\dots,N_\mathrm{p}-1   \;, \\
                              \quad &   e_k = r_k - Cz_k                                        \;, \\
                              \quad &   u_{\rm min} \leq  u_k \leq u_{\rm max}                  \;, \\
\textnormal{parameters}       \quad &   z_0 = \Psi(y_t)                                         \;, \\
                              \quad &   r_k = \textnormal{given}  \;,  \quad k =0,1,\dots,N_\mathrm{p}    \;, 
\end{split}
\end{equation}
where $u_{\rm min/max}$ are bounds on inputs.
After the optimal sequence $u_k$ is found, only the first element $u_0$ is used, new measurement is obtained, and the process repeats in next time step $t+1$. 

The loss function $J(\cdot)$ in~\eqref{eq:MPC_basic_opt_problem} has a standard quadratic form for an MPC tracking problem
\begin{equation}\label{eq::QP_cost}
J\left(z_k, u_k\right) = \frac{1}{2}e_{N_\mathrm{p}}^\top S e_{N_\mathrm{p}} + \frac{1}{2}\sum_{k = 0}^{{N_\mathrm{p}}-1}\left(e_k^\top Q e_k + u_k^\top R u_k\right) \;. 
\end{equation}
Matrices $Q\succeq 0$ and $S\succeq 0$ penalize the error inside and the end of the prediction horizon, respectively, while $R \succeq 0$ penalizes the input.

Furthermore, to reduce computational complexity, and due to computational considerations (see Remark~\ref{rem:dense_formulation}), we rewrite the quadratic problem~\eqref{eq:MPC_basic_opt_problem} into a \textit{dense} and \textit{delta input} formulation.
For the sake of brevity, we omit the full derivation and present only the resulting formulas; see~\cite{borrelli_predictive_2017} or other classical textbooks on MPC for full explanation.
Let 
\begin{equation}
  \begin{split}
    \Delta u_k &= u_{k} - u_{k-1} \;, \\
    \Delta U &= \left[ \Delta u_0^\top , \Delta u_1^\top, \ldots, \Delta u_{N_\mathrm{p} - 1}^\top  \right]^\top  \;,\\
    r &= [r_0^\top, r_1^\top, \ldots, r_{N_\mathrm{p} - 1}^\top]^\top  \;.
  \end{split}
\end{equation}
The final quadratic problem solved by MPC at each time step $t$ is
\begin{equation}\label{eq::KMPC_dense_opt}
\begin{split}
\min_{\Delta U} \quad  &  \frac{1}{2}\Delta U^\top H\Delta U + \left[ z_0^\top r^\top\right]F^\top\Delta U \;, \\
\text{ s.t.}    \quad  &  b_\mathrm{min} \leq G \Delta U \leq b_\mathrm{max}  \;,\\
\end{split} 
\end{equation} 
where the matrices $H$, $F$, $G$, and vectors $b_\mathrm{min}, b_\mathrm{max}$ can be derived from the state ($A,B,C$) and cost ($Q,R$) matrices.



\begin{remark}\label{rem:dense_formulation}
  We use the dense formulation so the dimensionality of the optimization problem~\eqref{eq::KMPC_dense_opt} does not scale with the number of observables (the size of the lifted state $z$).
  The delta input formulation allows achieving offset-free control in case of external disturbances acting on the system. 
\end{remark}

\section{Experiments}

\begin{figure} [tb]
\begin{subfigure}{.23\textwidth}
  \centering
  \includegraphics[width=.7\linewidth]{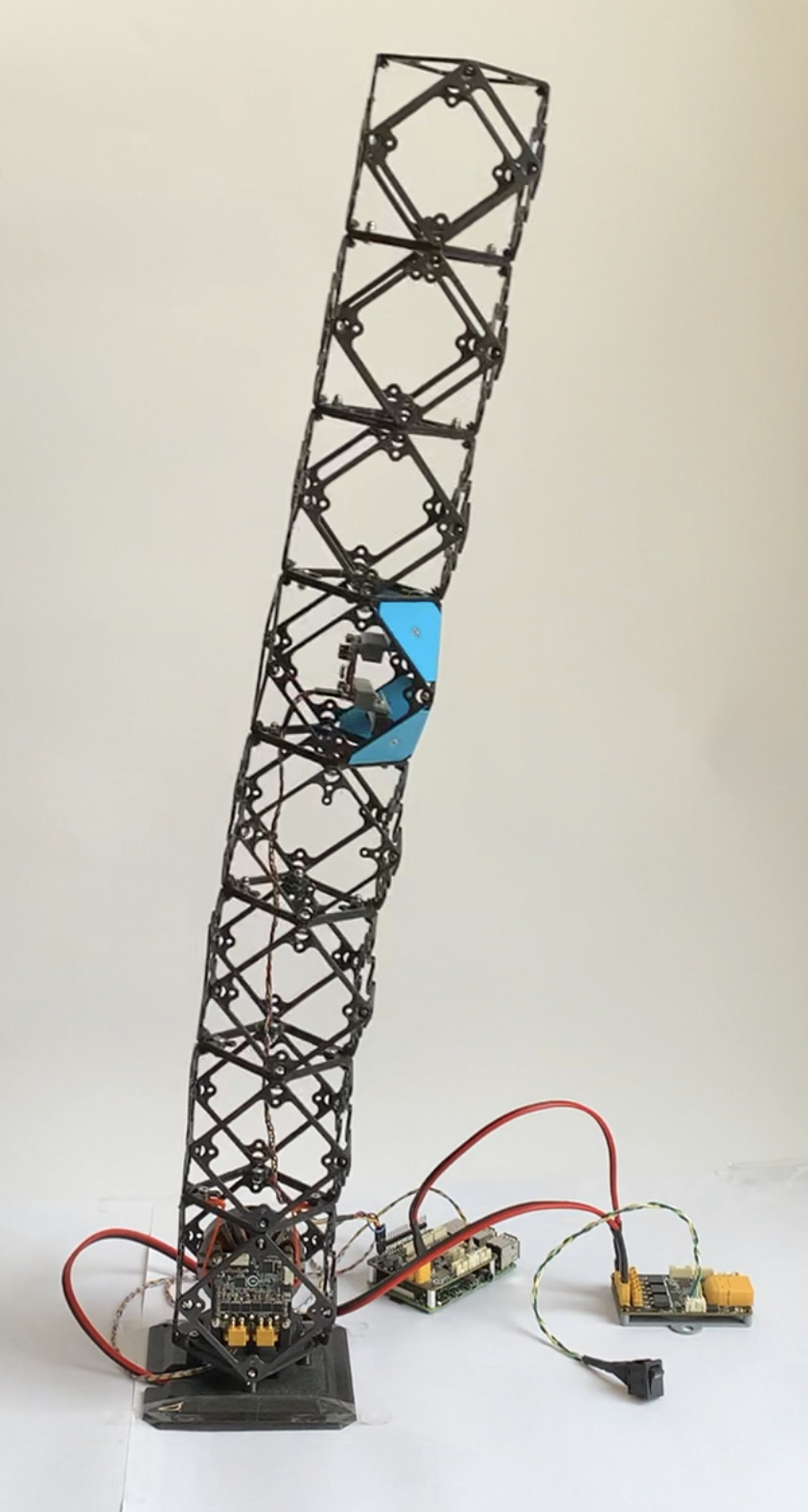}
  \caption{}\label{fig::VT_real_photo}
\end{subfigure}%
\begin{subfigure}{.23\textwidth}
  \centering
  \includegraphics[width=.85\linewidth]{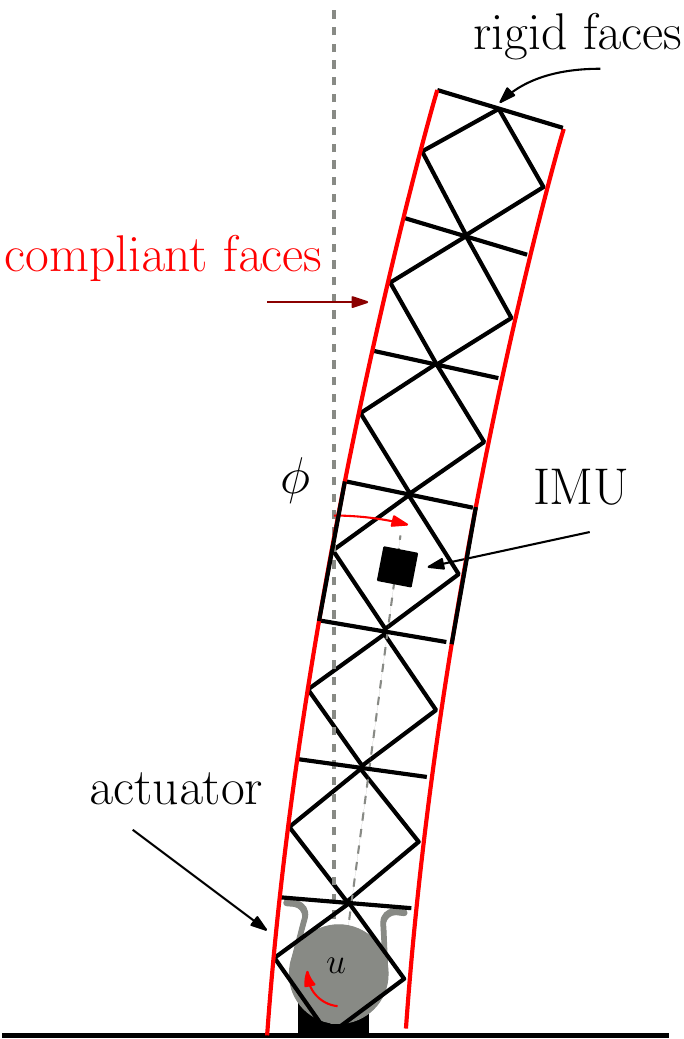}
  \caption{}\label{fig::VT_scheme}
\end{subfigure}
\caption[The Voxel Tower]{The Voxel Tower: (\subref{fig::VT_real_photo}) a photo of the real prototype;
    (\subref{fig::VT_scheme}) a graphic of the Voxel Tower system with its individual parts.}%
  \label{fig::VoxelTower}
\end{figure}

In this section, we present experiments with a lattice structure built from the voxels, the \textit{Voxel tower}.
The Voxel Tower is chosen as the experimental structure as it is a basic geometry applicable to a variety of different application, such as soft-robotic arms, limbs, or smart architecture. 
The structure, together with the scheme of its architecture, is depicted in Fig.~\ref{fig::VoxelTower}.
The tower's motion is controlled by the actuated voxel at the bottom, and the movement is measured by sensing voxel mounted halfway up the tower.

The structure consist of eight vertically connected voxels of two types.
The structure is primarily composed of semi-compliant voxels that enable tilting in one axis, while being stiff in the other. 
We additionally used one rigid voxel that serves as the housing for the sensor.



\subsection{Control Synthesis}

To identify the predictor~\eqref{eq:linear_predictor}, we gathered the inputs and outputs from the real system.
In total, we acquired ${p=\num{4e4}}$ measurements points from six trajectories.
Each trajectory consists of supplying a different open-loop input and measuring the tilt $\phi$ and tilt speed $\dot{\phi}$.

As observables, we used \textit{delay-embeddings} (time-shifted copies) of the measured output. 
The use of delay embeddings has been documented and justified in~\cite{tu_dynamic_2014}. 
For the stabilization task~\ref{sec::Stabilisation} we used $s=2$ delay embeddings, while the tracking predictor yielded better results with $s=3$. 
Upon introducing $\psi = \left[\phi, \dot{\phi}, \mathcal{D}u\right]$, where $\mathcal{D}^nb_k = b_{k-n}$ is the delay operator, we can write the particular lifting functions as
\begin{align}
  \Psi_{\rm stab.}(\psi) & =  [\mathcal{D}^2\psi,\mathcal{D}\psi, \psi]^\top \;, \\
  \Psi_{\rm trac.}(\psi) &=  [\mathcal{D}^3\psi, \mathcal{D}^2\psi,\mathcal{D}\psi, \psi]^\top\;.
\end{align}

The main tuning parameters for the design of the stabilizing LQR controller are the values of matrices $Q$ and $R$ in~\eqref{eq::LQR_cost}. 
Since penalizing delayed states or inputs is not meaningful, only the values of $Q$ corresponding with the system's original states are non-zero and 
chosen to obtain the desired performance.

For the design of the KMPC controller, we still have the state and input penalty matrices $Q$ and $R$ at hand, but on top of that, there are two extra parameters, that we can use -- the prediction horizon length $N_\mathrm{p}$, and the final state penalty matrix $S$. 
We set the prediction horizon $N_\mathrm{p} = 10$, and through a suitable choice of the matrix $C$ in~\eqref{eq:MPC_basic_opt_problem}, we can again penalize only the original states $\phi$ and $\dot{\phi}$. 
The corresponding values in the $Q$ and $S$ matrices that result in the desired performance were chosen.  

The KMPC gives us the option to bound the control action. 
We found the limit on the applied torque to be ${\pm \SI{0.5}{\newton\meter}}$. 
This limit ensured the torque would not damage the actuated voxel, while still deforming the voxel enough to induce motion into the structure.

\subsection{Stabilization} \label{sec::Stabilisation}

\begin{figure}
  \centering
    \begin{subfigure}{.45\textwidth}
    \centering
    \includegraphics[width=0.97\linewidth]{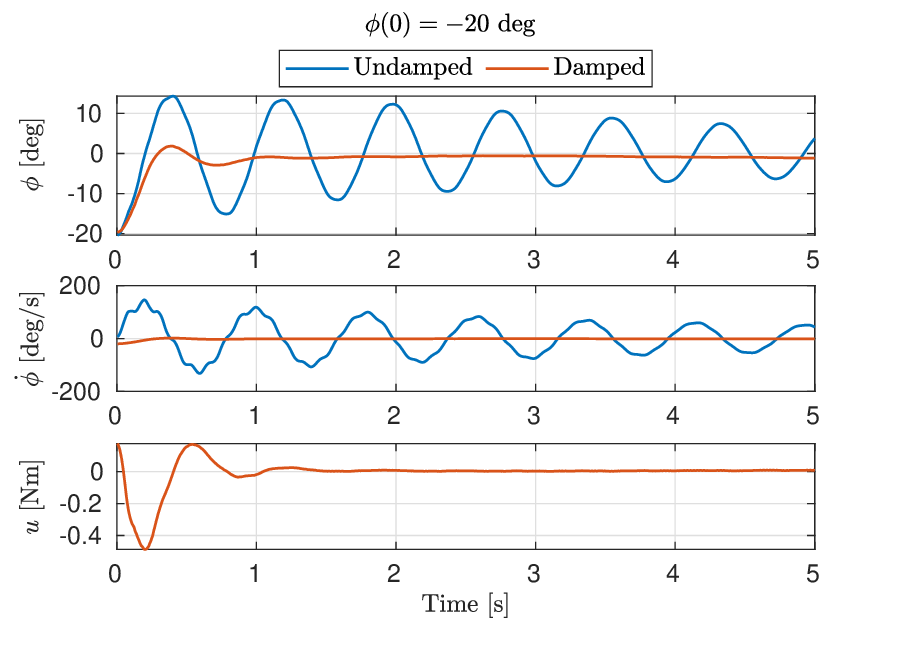}
    \caption{}\label{fig:nz_tilt}
    \end{subfigure}	

    \begin{subfigure}{.15\textwidth}
    \centering
    \includegraphics[width=.97\linewidth]{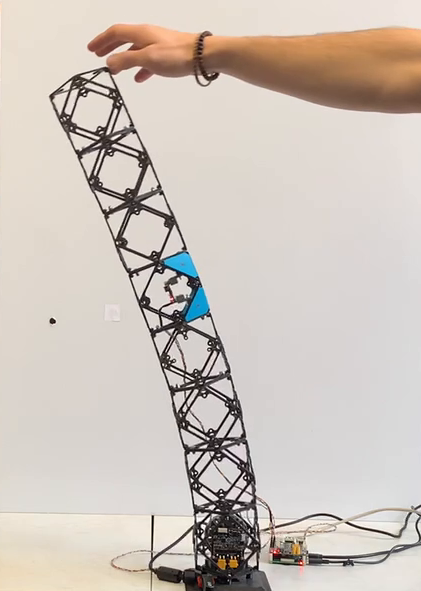}
    \caption{}\label{fig::step1}
    \end{subfigure}%
    \begin{subfigure}{.15\textwidth}
    \centering
    \includegraphics[width=.97\linewidth]{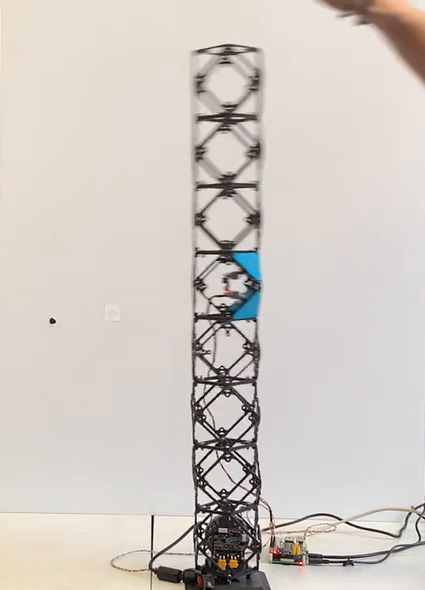}
    \caption{}\label{fig::step2}
    \end{subfigure}%
    \begin{subfigure}{.15\textwidth}
    \centering 
    \includegraphics[width=.97\linewidth]{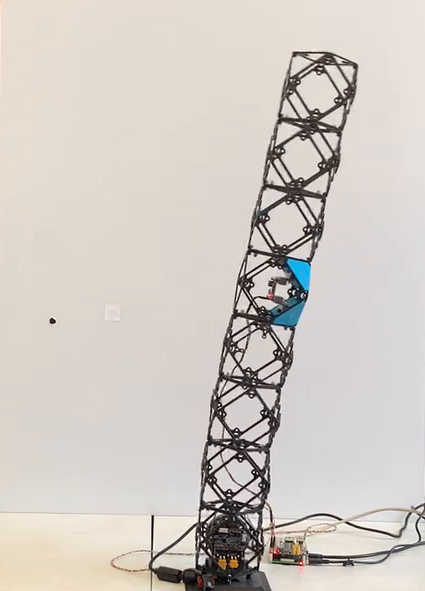}
    \caption{}\label{fig::step3}
    \end{subfigure}

    \begin{subfigure}{.15\textwidth}
    \centering
    \includegraphics[width=.97\linewidth]{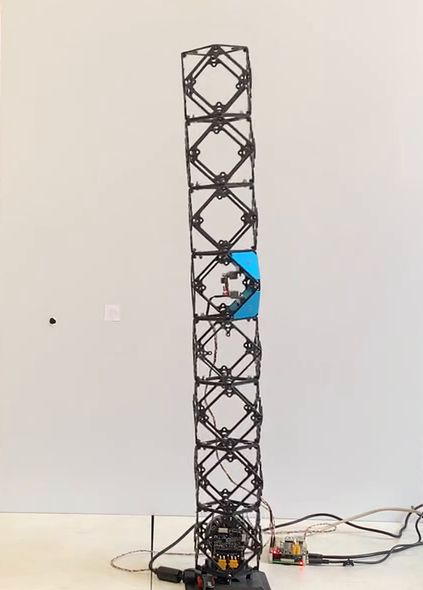}
    \caption{}\label{fig::step4}
    \end{subfigure}%
    \begin{subfigure}{.15\textwidth}
    \centering
    \includegraphics[width=.97\linewidth]{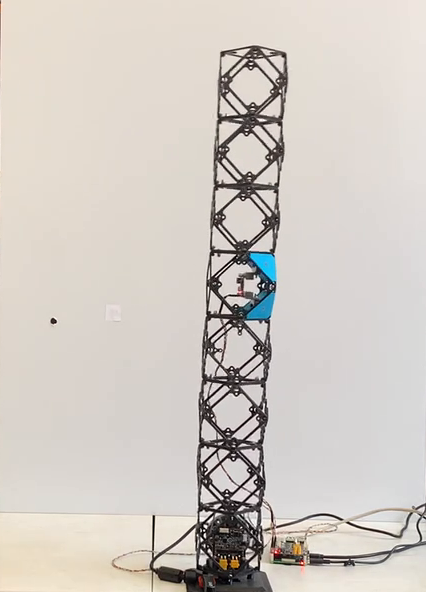}
    \caption{}\label{fig::step5}
    \end{subfigure}%
    \begin{subfigure}{.15\textwidth}
    \centering 
    \includegraphics[width=.97\linewidth]{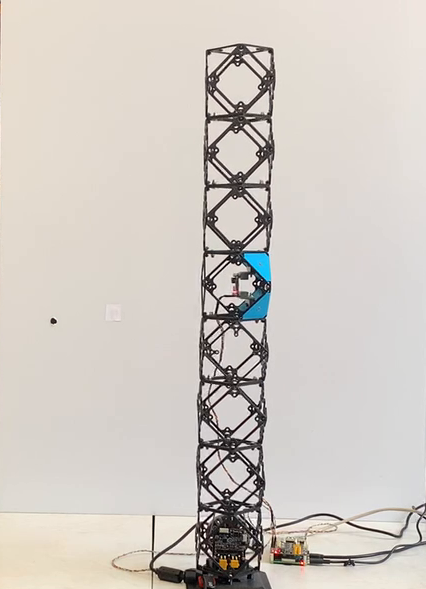}
    \caption{}\label{fig::step6}
    \end{subfigure}
    \caption{(\subref{fig:nz_tilt}) Response to the initial deviation $\phi_0 = \SI{-20}{\degree}$ from vertical position
      Snapshots of the experiment with active damping:
    (\subref{fig::step1}) $t = \SI{0}{\second}$;
    (\subref{fig::step2}) $t = \SI{0.2}{\second}$;
    (\subref{fig::step3}) $t = \SI{0.5}{\second}$;
    (\subref{fig::step4}) $t = \SI{0.8}{\second}$;
    (\subref{fig::step5}) $t = \SI{1.0}{\second}$;
    (\subref{fig::step6}) $t = \SI{1.4}{\second}$.}
    \label{fig::AV_experiment_snapshots}
\end{figure}

In the first testing scenario, presented in Fig.~\ref{fig::AV_experiment_snapshots}, we tilted the tower into an initial angle deviation: $\phi_0 = \SI{-20}{\degree}$ and compared the response in the controlled system to the uncontrolled system.
The figure includes snapshots from the experiment.
Even though the uncontrolled structure would also eventually stabilize due to damping, the controlled structure is able to stabilize much faster. 

\begin{figure}
  \begin{subfigure}{.45\textwidth}
  \centering
  \includegraphics[width=\linewidth]{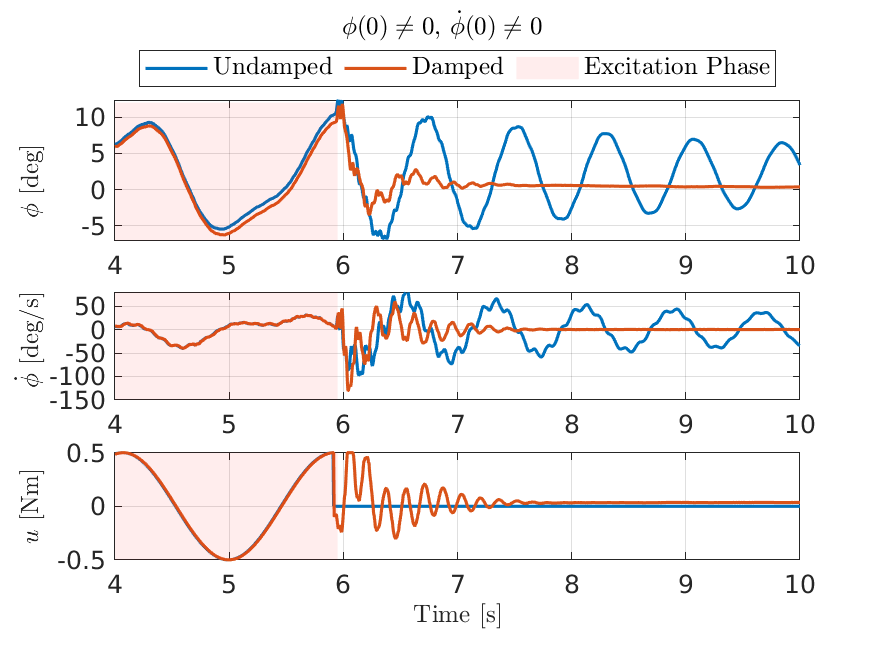}
  \caption{}\label{fig::VT_nzv_1}
  \end{subfigure}
  ~
  \begin{subfigure}{.45\textwidth}
  \centering
  \includegraphics[width=\linewidth]{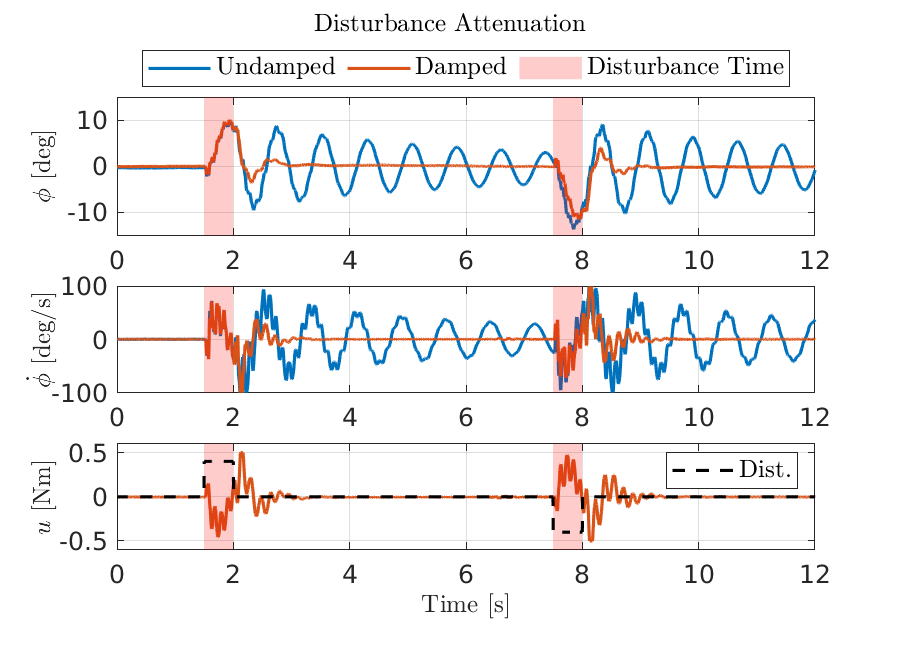}
  \caption{}\label{fig::VT_dist_rej}
  \end{subfigure}
   \caption{Two experiments of stabilization
  }
  \label{fig::nzv_and_dist_rej}
\end{figure}

In the second scenario, presented in Fig.~\ref{fig::VT_nzv_1}, we tested the response in the case where neither of the state variables is zero.
The figure compares the behavior of the controlled and uncontrolled systems.  
We first applied a precomputed sine wave torque excitation to the system (red area) to initiate movement.
After 6 seconds, when the tower was already in motion, we switched to the damping control. 
Again, we can observe that the controlled system significantly outperforms the uncontrolled system in terms of settling time.

In the third scenario, presented in Fig.~\ref{fig::VT_dist_rej}, we applied two $\SI{0.5}{\second}$ force pulses, simulating disturbances. 
The disturbance signal is shown in the bottom plot as a black dashed line.
During the time of the disturbance (red areas), the motor acts as the disturbance. 
The figure also shows the response of the tower to the disturbance, comparing the behavior of the uncontrolled system with the controlled system. 
The measured results show that the designed controller is able to successfully dampen even higher-frequency oscillations caused by the abrupt disturbance.

\subsection{Tracking} \label{sec::Tracking}

\begin{figure} [tb]
  \centering
  \begin{subfigure}{.45\textwidth}
    \centering
    \includegraphics[width=1\linewidth]{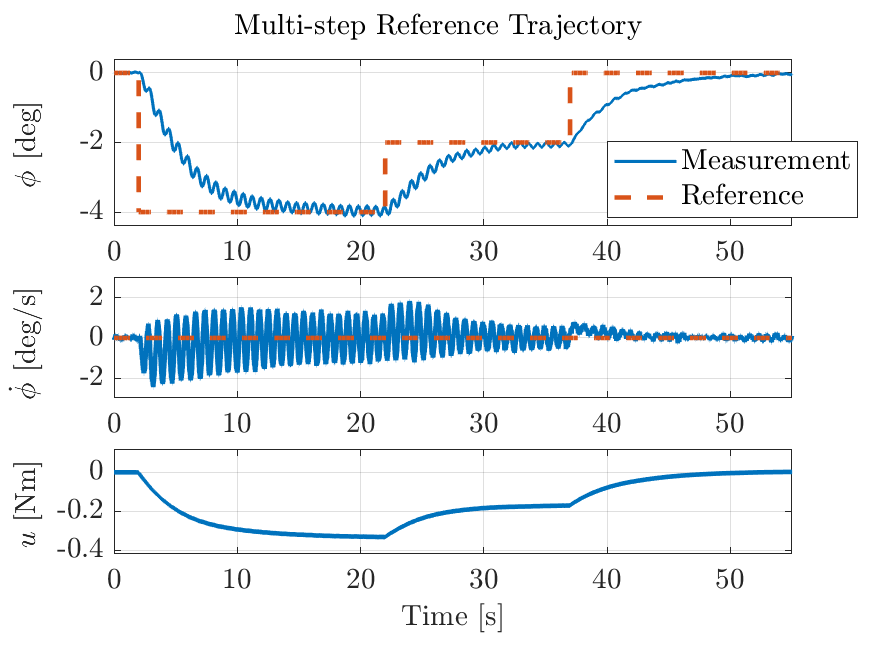}
    \caption{}\label{fig::VT_tracking_seq_steps1}
  \end{subfigure}
  \begin{subfigure}{.45\textwidth}
    \centering
    \includegraphics[width=1\linewidth]{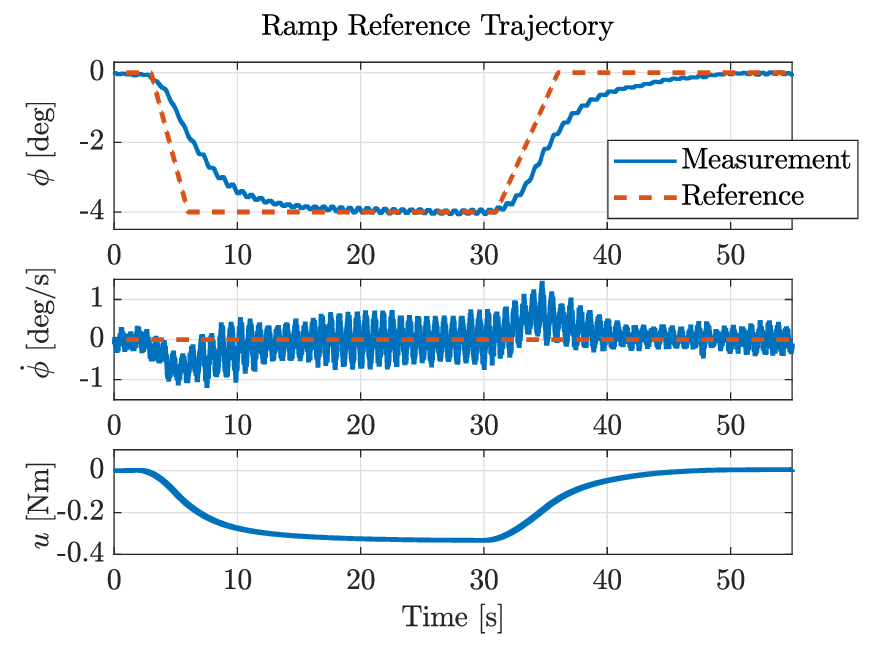}
    \caption{}\label{fig::VT_tracking_ramp}
  \end{subfigure}  
  \caption{Two experiments of reference tracking}\label{fig::VT_step_seq1_tracking}
\end{figure}

To test the KMPC controller, we generated two reference trajectories.
First, we required multiple consecutive step changes in the tilt angle. 
The second type of tested trajectories consist of gradual changes in the tilt angle. 
In both experiments, the reference velocity remained zero. 
This choice was made considering that the tower was intended to remain stationary or move slowly.
We present the measured responses in Fig.~\ref{fig::VT_step_seq1_tracking}.
The outcomes of both experiment confirm that the KMPC is able to track the desired references and further confirms the efficiency of the developed method.

\section{Conclusion} \label{sec::Conclusion}

This work introduced sensing and feedback control into lattice structures made from 3D printed cuboctahedron voxels. 
We presented a novel means for actuation, the \textit{actuated} voxel, which uses local deformations to actuate a voxel structure.
We build a structure from these digital materials-- the Voxel Tower-- and considered the problem of stabilization, disturbance attenuation, and reference tracking. 
To solve these tasks, we used the EDMD algorithm to obtain a linear predictor and implemented two regulators, the LQR and the Koopman MPC, and demonstrated their performance on the physical system.
The presented work expanded the capabilities of mechanical lattice structures beyond open-loop control and can serve as a stepping stone toward the utilization of digital structures in complex applications that demand more advanced control techniques.
Future work will address controlling systems with more complex structure.

\bibliography{biblio}

\end{document}